\documentclass{article}

\usepackage[final]{nips_2017}

\usepackage[utf8]{inputenc} 
\usepackage[T1]{fontenc}    
\usepackage{hyperref}       
\usepackage{url}            
\usepackage{booktabs}       
\usepackage{amsfonts}       
\usepackage{nicefrac}       
\usepackage{microtype}      
\usepackage{amsmath}
\usepackage[pdftex]{graphicx}
\usepackage{rotating}

\title{MANTIS: Model-Augmented Neural neTwork with Incoherent \textit{k}-space Sampling for efficient MR T2 mapping}

\author{Fang Liu \\
Department of Radiology\\
University of Wisconsin-Madison\\
Madison, Wisconsin USA, 53705-2275 \\
  \texttt{leoliuf@gmail.com} \\
  \And
Li Feng\\
Department of Medical Physics\\
Memorial Sloan Kettering Cancer Center \\
New York, NY USA, 10065 \\
\texttt{fengl@mskcc.org} \\
  \And
Richard Kijowski\\
Department of Radiology\\
University of Wisconsin-Madison\\
Madison, Wisconsin USA, 53705-2275 \\
\texttt{RKijowski@uwhealth.org}
}

\begin{document}

\maketitle

\begin{abstract}
Purpose: To develop and evaluate a novel deep learning-based image reconstruction approach called MANTIS (Model-Augmented Neural neTwork with Incoherent \textit{k}-space Sampling) for efficient MR T2 mapping.
Methods: MANTIS combines end-to-end convolutional neural network (CNN) mapping, incoherent \textit{k}-space undersampling, and a physical model as a synergistic framework. The CNN mapping directly converts a series of undersampled images directly into MR T2 maps using supervised training. Signal model fidelity is enforced by adding an additional pathway between the undersampled \textit{k}-space and estimated T2 maps to ensure that the T2 maps produced synthesized \textit{k}-space consistent with the acquired undersampling measurements. The MANTIS framework was evaluated on T2 mapping of the knee at different acceleration rates and was compared with standard CNN mapping and conventional sparsity-based iterative reconstruction approaches. Global quantitative assessment and regional T2 analysis for the cartilage and meniscus were performed to demonstrate the reconstruction performance of MANTIS.
Results: MANTIS achieved high quality T2 mapping at both moderate (R=5) and high (R=8) acceleration rates. Compared to conventional reconstruction approaches that exploited image sparsity, MANTIS yielded lower errors (normalized root mean square error of 6.1\% for R=5 and 7.1\% for R=8) and higher similarity (structural similarity index of 86.2\% at R=5 and 82.1\% at R=8) with respect to the reference in the T2 estimation. MANTIS also achieved superior performance compared to standard CNN mapping.
Conclusion: The MANTIS framework, with a combination of end-to-end CNN mapping, signal model-augmented data consistency, and incoherent \textit{k}-space sampling, is a promising approach for efficient estimation of T2 maps.

\end{abstract}

\section{INTRODUCTION}
Quantitative mapping of magnetic resonance (MR) parameters, such as the spin-lattice relaxation time (T1) and the spin-spin relaxation time (T2), have been shown as valuable methods for improved assessment of a range of diseases. Compared to conventional MR imaging, parameter mapping can provide increased sensitivity to different pathologies with more specific information on tissue composition and microstructure. Standard approaches for estimation of MR parameters usually require repeated acquisitions of datasets with different imaging parameters, for example, multiple echo times (TEs) for T2 mapping and multiple flip angles (FAs) or inversion recovery times (TIs) for T1 mapping. The corresponding parameter values can then be generated by fitting the acquired images to a physical model on a pixel-by-pixel basis, yielding a parameter map. Due to the need to image an anatomic structure multiple times, parameter mapping usually requires long scan times compared to conventional imaging, limiting its widespread clinical use. Therefore, accelerated parameter mapping is highly-desirable and remains a topic of great interest in the MR research community.

To allow for rapid parameter mapping, many approaches can be applied to accelerate data acquisitions, such as parallel imaging utilizing multi-coil sensitivities (1–3), compressed sensing exploiting image sparsity (4), or a combination of both (5,6). Data acquisitions can also be further accelerated by reconstructing undersampled dynamic images in a joint spatial and parametric space (x-p space) to explore spatial-temporal correlations (7), or by additionally incorporate a model into the reconstruction process (8–12). Indeed, the high correlations presented in the parametric dimension offer an efficient way for exploring signal models as prior knowledge for image reconstruction. For example, several methods have been proposed to explore temporal correlations along the parameter dimension to highly accelerate data acquisitions (11–13).  Other methods aim to constrain signal evolution in the parametric dimension using an analytical model for further improved reconstruction performance (8,14,15). More recently, MR fingerprinting (MRF) (16), a technique which takes advantage of incoherent signal acquisition schemes in combination with pattern recognition of numerically simulated signal dictionary, has provided fast and artifact-insensitive parameter mapping in several applications (17).

There has been much recent interest in applying deep learning to a wide variety of MR imaging applications.  Deep learning methods have been successful used for image classification, tissue segmentation, object recognition, and image registration.  There have also been recent works describing the use of deep learning in image reconstruction with promising initial results. For example, Hammernik et al have recently proposed a generalized compressed sensing framework using a variational network for accelerated imaging of the knee (18). This approach aims to learn an optimal regularization function and reconstruction setting for improved reconstruction performance. Other approaches attempt to extend the compressed sensing framework using different deep learning architectures and have achieved success for image reconstruction (19–21). Meanwhile, various approaches have also been proposed to directly remove aliasing artifacts from undersampled images using a direct end-to-end convolutional neural network (CNN) mapping (22–27).  Zhu et al have also proposed an approach called AUTOMAP to directly estimate artifact-free images from undersampled \textit{k}-space using the so-called domain-transform learning, which has demonstrated the feasibility of learning mutually correlated information from multiple manifolds (28). While these deep learning methods have focused on highly efficient image reconstruction for conventional static MR imaging, applications of deep learning for dynamic imaging and in particular accelerated parameter mapping have been limited (29–31).

The purpose of this work was to develop and evaluate a novel deep learning-based reconstruction framework called Model-Augmented Neural neTwork with Incoherent \textit{k}-space Sampling (MANTIS) for efficient T2 mapping. Our approach combines end-to-end CNN mapping with \textit{k}-space consistency using the concept of cyclic loss (20,32) to further enforce data and model fidelity. Incoherent \textit{k}-space sampling is used to improve reconstruction performance. A physical model is incorporated into the proposed framework, so that the parameter maps can be efficiently estimated directly from undersampled images. The performance of MANTIS was demonstrated for T2 mapping of the knee joint

\section{THEORY}
This section describes details about the MANTIS framework tailored for T2 mapping. However, such a framework can also be modified to estimate other quantitative MR parameters with corresponding signal models.

\subsection{Model-Based Reconstruction for Accelerated T2 Mapping}

In model-based reconstruction for accelerated T2 mapping, an image needs to be estimated from a subset of \textit{k}-space measurements for the $j^{th}$ echo first, which can be written as: 

\begin{equation} \label{eq1}
{d_j} = E{i_j} + \varepsilon
\end{equation}

Here $\varepsilon$ is the complex Gaussian noise in the measurements (33), $d_j$ is the undersampled \textit{k}-space data, and $i_j$ is the corresponding image with a size of $n_x \times n_y$ to be reconstructed satisfying the T2 signal behavior at the $j^{th}$ echo time ($TE_j$) as: 

\begin{equation} \label{eq2}
{i_j} = {S_j}\left( {{{\rm{I}}_{\rm{0}}},{{\rm{T}}_{\rm{2}}}} \right) = {{\rm{I}}_0} \cdot {e^{ - {{T{E_j}} \mathord{\left/
 {\vphantom {{T{E_j}} {{{\rm{T}}_{\rm{2}}}}}} \right.
 \kern-\nulldelimiterspace} {{{\rm{T}}_{\rm{2}}}}}}}
\end{equation}

where $I_0$ and $T_2$ represent the proton density image and associated T2 map, respectively. The encoding matrix $E$ can be expanded as: 

\begin{equation} \label{eq3}
E = MF
\end{equation}

where $F$ is an encoding operator performing Fourier Transform and $M$ is an undersampling pattern selecting desired \textit{k}-space measurements. A signal-to-noise ratio (SNR) optimized reconstruction of Eq.\ref{eq1} can be accomplished by minimizing the following least squares errors: 

\begin{equation} \label{eq4}
{{\rm{\tilde I}}_{\rm{0}}},{{\rm{\tilde T}}_{\rm{2}}} = \arg \;\mathop {\min }\limits_{{{\rm{I}}_{\rm{0}}},{{\rm{T}}_{\rm{2}}}} \sum\limits_{j = 1}^t {\left\| {E{S_j}\left( {{{\rm{I}}_{\rm{0}}},{{\rm{T}}_{\rm{2}}}} \right) - {d_j}} \right\|_2^2}
\end{equation}

where ${\left\|  \cdot  \right\|_2}$ denotes the $l_2$ norm and $t$ is the total number of echoes. Since the system of Eq.\ref{eq4} can be poorly conditioned at high acceleration rates, the reconstruction performance can be improved by minimizing a cost functional that includes additional regularization terms on the to-be-reconstructed proton density image and T2 map: 

\begin{equation} \label{eq5}
{{\rm{\tilde I}}_{\rm{0}}},{{\rm{\tilde T}}_{\rm{2}}} = \arg \;\mathop {\min }\limits_{{{\rm{I}}_{\rm{0}}},{{\rm{T}}_{\rm{2}}}} \left({\frac{1}{2}\sum\limits_{j = 1}^t {\left\| {E{S_j}\left( {{{\rm{I}}_{\rm{0}}},{{\rm{T}}_{\rm{2}}}} \right) - {d_j}} \right\|_2^2}
 + \sum\limits_k {{\lambda _k}{R_k}({{\rm{I}}_{\rm{0}}},{{\rm{T}}_{\rm{2}}})}}\right)
\end{equation}

The regularization penalty ${R_k}({{\rm{I}}_{\rm{0}}},{{\rm{T}}_{\rm{2}}})$ can be selected based on prior knowledge or assumptions about the model considering desired parameters (8–12), with a weighting parameter  ${\lambda _k}$ controlling the balance between the data fidelity (the left term) and the regularization (the right terms). Eq.\ref{eq5} describes the generalized model-based reconstruction framework that can be implemented for accelerated quantitative T2 mapping (7,8).

\subsection{End-to-End CNN Mapping}

End-to-end mapping using CNN as a nonlinear mapping function has been shown to be quite successful in recent applications of domain-to-domain translation. The concept behind the mapping method is to use CNN to learn spatial correlations and contrast relationships between input datasets and desirable outputs. Such a network structure has been shown to capable of mapping from one image domain to another image domain representing discrete tissue classes (e.g. image segmentation (34–36)), from artifact or noise corrupted images to artifact-free images (e.g. image restoration and reconstruction (22–27)), from one image contrast to a different image contrast (e.g. image synthesis (37–41)), and from \textit{k}-space directly to image space (e.g. domain-transform   learning (28)). In the current study, end-to-end CNN mapping was performed from the undersampled image domain to the parameter domain, denoted as ${i_u} \to ({{\rm{\tilde I}}_{\rm{0}}},{{\rm{\tilde T}}_{\rm{2}}})$, where the undersampled image can be obtained from fully sampled image retrospectively using zero-filling reconstruction:

\begin{equation} \label{eq6}
{i_u} = {E^H}Fi
\end{equation}

Here, $E^H$ represents the Hermitian transpose of the encoding matrix $E$ shown in Eq.\ref{eq3}.

To perform parameter mapping using CNN, a deep learning framework is designed that regularizes the estimation by satisfying the constraints of prior knowledge in the training process (i.e. regularization by prior knowledge in the training datasets). Since no particular assumptions are made in the training process, the prior knowledge originates from certain latent features that are learned from the datasets during network training (18,23,30). In another word, the network aims to learn a model that will map the undersampled images directly to the parameter maps that will be estimated. The end-to-end CNN mapping from the undersampled images (in $domain({i_u})$) to the T2 parameter maps can be expressed as: 

\begin{equation} \label{eq7}
{R_{cnn}}({{\rm{I}}_{\rm{0}}},{{\rm{T}}_{\rm{2}}}) = {{\rm E}_{{i_u} \to domain({i_u})}}\left[ {{{\left\| {C({i_u}|\theta ) - ({{\rm{I}}_{\rm{0}}}{\rm{,}}{{\rm{T}}_{\rm{2}}})} \right\|}_2}} \right]
\end{equation}

Here, $C({i_u}|\theta ):{i_u} \to ({{\rm{\tilde I}}_{\rm{0}}},{{\rm{\tilde T}}_{\rm{2}}})$ is a mapping function conditioned on network parameter $\theta$, and the ${{\rm E}_{{i_u} \to domain({i_u})}}\left[  \cdot  \right]$ is an expectation operator of a probability function given $i_u$ belongs to a training dataset $domain({i_u})$. As in most deep learning works, the $l_2$ norm is typically selected as a loss function to ensure that the mapping is accurate(18,22–26). 

\subsection{MANTIS: Extending End-to-End CNN Mapping with Model-Consistency}
Similar to prior studies for CNN-based image reconstruction with a focus on data consistency (20,23), Eq.\ref{eq7} is inserted into Eq.\ref{eq5} to replace the original regularization term so that a new objective function (Eq.\ref{eq8}) can be formulated, in which the CNN-based mapping serves as a regularization penalty term (right term) to the data consistency term (left term). 

\begin{multline} \label{eq8}
\tilde \theta  = \arg \;\mathop {\min }\limits_\theta  \left( {\lambda _{data}}{{\rm E}_{{i_u} \to domain({i_u})}}\left[ {\sum\limits_{j = 1}^t {\left\| {E{S_j}(C({i_u}|\theta )) - {d_j}} \right\|_2^2} } \right]
\right.
\\
\left.
\vphantom{\left[ {\sum\limits_{j = 1}^t {\left\| {E{S_j}(C({i_u}|\theta )) - {d_j}} \right\|_2^2} } \right]} + {\lambda _{cnn}}{{\rm E}_{{i_u} \to domain({i_u})}}\left[ {{{\left\| {C({i_u}|\theta ) - ({{\rm{I}}_{\rm{0}}}{\rm{,}}{{\rm{T}}_{\rm{2}}})} \right\|}_2}} \right] \right)
\end{multline}

Here, $\lambda_{data}$ and $\lambda_{cnn}$ are regularization parameters balancing the model fidelity and CNN mapping, respectively, assuming that there is a training database including the fully- sampled images $i$ and a pre-defined undersampling mask $M$.

\begin{figure}[h]
  \centering
  \includegraphics[width=0.8\linewidth]{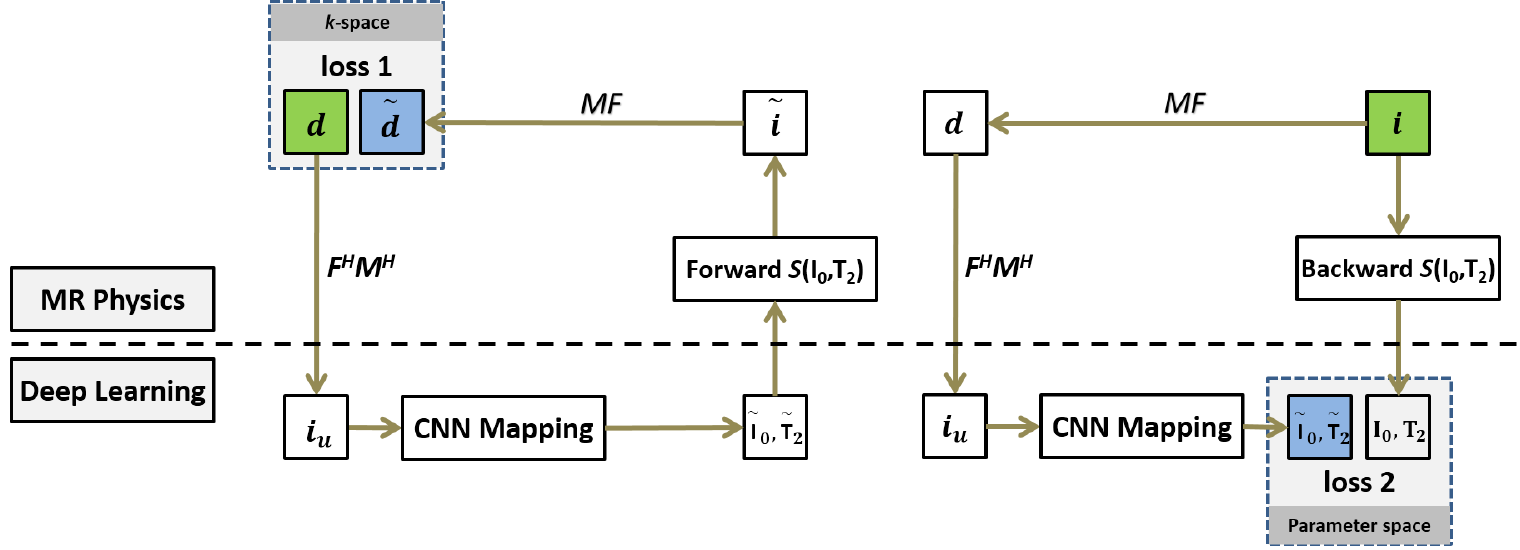}
  \caption{Illustration of the MANTIS framework, which features two loss components. The first loss term (loss 1) ensures that the reconstructed parameter maps from the CNN mapping produces synthetic undersampled \textit{k}-space data ($\tilde d$) matching the acquired \textit{k}-space measurements ($d$). The second loss term (loss 2) ensures that the undersampled images produces parameter maps (${{\rm{\tilde I}}_{\rm{0}}},{{\rm{\tilde T}}_{\rm{2}}}$) that are same as the reference parameter maps (${{\rm{I}}_{\rm{0}}},{{\rm{T}}_{\rm{2}}}$). The MANTIS framework considers both the data-driven deep learning component and signal model from the basic MR physics. The notation in this figure follows the main text description.}
\label{fig1}
\end{figure}

The training process of Eq.\ref{eq8} is equivalent to training two cyclic losses (20,32) as shown in the Figure\ref{fig1}. The first loss term (loss 1) ensures that the reconstructed parameter maps from CNN mapping produce undersampled \textit{k}-space data matching the acquired \textit{k}-space measurements. The second loss term (loss 2) ensures that the undersampled images produce the same parametric maps as the reference parameter maps (i.e. an objective in normal supervised learning). Note that Eq.\ref{eq8} is fundamentally different from Eq.\ref{eq5} with respect to the optimization target. While the conventional model-based reconstruction in Eq.\ref{eq5} is attempting to reconstruct each individual set $({{\rm{\tilde I}}_{\rm{0}}},{{\rm{\tilde T}}_{\rm{2}}})$ matching the acquired \textit{k}-space data, the new framework in Eq.\ref{eq8} is attempting to estimate a parameter set   conditioned on which the CNN mapping optimizes the estimation performance in the current training datasets. The data-consistency term (the left term in Eq.\ref{eq8}) further ensures that the estimation during the training process is correct in \textit{k}-space. Once the training process is completed, the estimated parameter set $\tilde \theta $ is fixed and it can be used to efficiently convert new undersampled images to their corresponding parameter maps $({{\rm{\tilde I}}_{\rm{0}}},{{\rm{\tilde T}}_{\rm{2}}})$  directly, formulated as: 

\begin{equation} \label{eq9}
{{\rm{\tilde I}}_{\rm{0}}},{{\rm{\tilde T}}_{\rm{2}}} = C({i_u}|\tilde \theta ),{i_u} \to domain({i_u})
\end{equation}

\section{METHODS}
\subsection{In-Vivo Image Datasets}

This retrospective study was performed in compliance with Health Insurance Portability and Accountability Act (HIPPA) regulations, with approval from our Institutional Review Board, and with a waiver of written informed consent. T2 mapping images of the knee acquired in 100 symptomatic patients (55 males and 45 females, mean age = 65 years) on a 3T scanner (Signa Excite Hdx, GE Healthcare, Waukesha, Wisconsin) equipped with an eight-channel phased-array extremity coil (InVivo, Orlando, Florida) were retrospectively collected for training a deep neural network.  All images were acquired using a multi-echo spin-echo T2 mapping sequence in sagittal orientation with the following imaging parameters: field of view (FOV) = 16$\times$16cm2, repetition time (TR) = 1500ms, echo times (TEs) = [7, 16, 25, 34, 43, 52, 62, 71] ms, flip angle = 90 degree, bandwidth = 122Hz/pixel, slice thickness = 3-3.2mm, number of slices = 18-20, and acquired image matrix = 320$\times$256 which was interpolated to 512$\times$512 after reconstruction. The images were reconstructed directly on the MR scanner and were saved as magnitude images into DICOM files after coil combination. These image datasets were treated as “fully sampled” reference used in the network training. To evaluate the trained network, T2 mapping images of the knee were acquired on additional 10 symptomatic patients (7 males and 3 females, mean age = 63 years) and one healthy volunteer using the same T2 mapping sequence with matching imaging parameters. For the patient scans, images were reconstructed directly on the MR scanner and were saved as separate magnitude and phase images into DICOM files. To demonstrate the generality for processing raw images, the multicoil \textit{k}-space data were saved directly from the scanner and were used for evaluation for the volunteer scan. 

\subsection{Implementation of Neural Network}

\begin{figure}[h]
  \centering
  \includegraphics[width=0.8\linewidth]{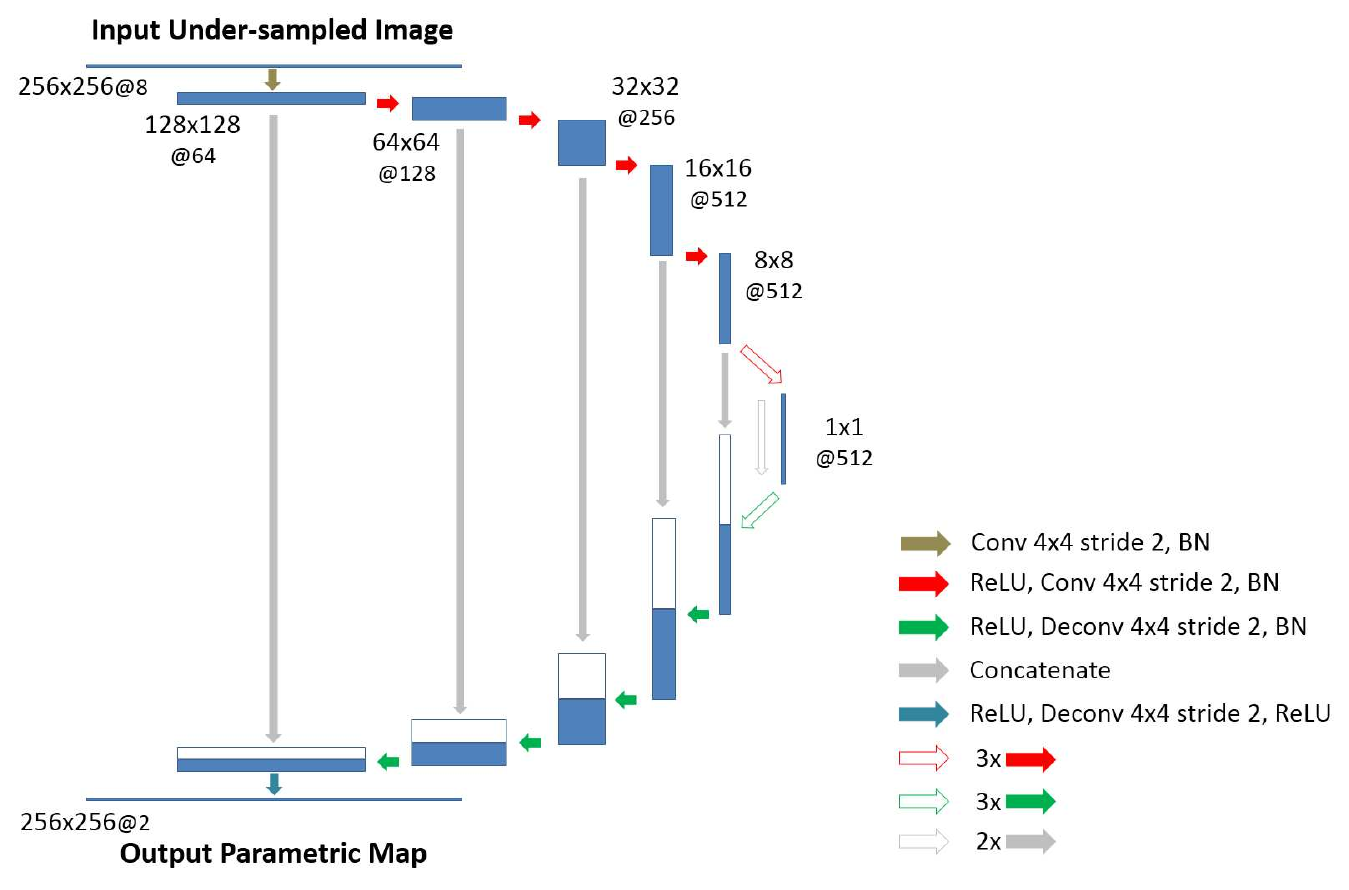}
  \caption{Illustration of the U-Net implemented in MANTIS for end-to-end CNN mapping. The U-Net structure consists of an encoder network and a decoder network with multiple shortcut connection (e.g. concatenation) between them to enhance mapping performance. The abbreviations for the CNN layers include: BN for Batch Normalization, ReLU for Rectified Linear Unit activation, Conv for 2D convolution, and Deconv for 2D deconvolution. The parameters for the convolution layers are labelled in the figure as image size @ the number of 2D filters.}
\label{fig2}
\end{figure}

A standard U-Net architecture (42) was adapted for the implementation of MANTIS for mapping the undersampled image datasets directly to corresponding T2 maps. As illustrated in Figure\ref{fig2}, the U-Net structure consists of an encoder network and a decoder network with multiple shortcut connections between them. The encoder network is used to achieve efficient data compression while probing robust and spatial invariant image features of the input images.  The decoder network, designed with a mirrored structure, is applied for restoring image features and increasing image resolution using the output of the encoder network. Multiple shortcut connections are incorporated to concatenate entire feature maps from the encoder to the decoder to enhance mapping performance. This type of network structure has shown promising results for image-to-image translation in many recent studies (24,35,39,42).

\subsection{Network Training}

The images are concatenated from all the eight echoes together, so that the network input has a total of 8 channels, as shown in Figure 2. From the 100 patient image datasets in the training group, 90 patient datasets were randomly selected for network training, while the remaining 10 patient datasets were used for validation during the training process to select the best network model. It should be noted that the 10 patient datasets used for network training validation were different from the 10 patient datasets used for final evaluation of the best network model.

The input of the framework included the desired undersampling masks for each image (described in the following subsection) and the undersampled multi-echo images, which were generated by multiplying the reference fully sampled \textit{k}-space data with the undersampling masks. The output of the network was a T2 map and a proton density (I0) image directly estimated from the undersampled multi-echo images. The estimated T2 map and proton density image were then validated against the reference T2 map and proton density images obtained by fitting the reference multi-echo images to an exponential model (Eq.\ref{eq2}) using a nonlinear least squares algorithm, as shown in Ref. (43). Such an iterative procedure continued to refine the network until it converged.

Due to GPU memory limitations, every input two-dimensional (2D) image was down-sampled to a matrix size of 256$\times$256 using bilinear interpolation. This step was performed due to the limited GPU memory in the server and could be avoided if sufficient memory was available. Image normalization was performed using data from each patient by dividing the complex-valued image by its maximum magnitude signal intensity. During the training of the network, the network weights were initialized using the initialization scheme suggested by He et al. (44) and were updated using an adaptive gradient decent optimization (ADAM) algorithm (45) with a fixed learning rate of 0.0002.  The network was trained in a mini-batch manner with three image slices in a single mini-batch. A total iteration steps corresponding to 200 epochs of the training dataset were carried out for the training, and the best model was selected as the one that provided the lowest loss value in the validation datasets. The parameters in the objective function (Eq.\ref{eq8}) were empirically selected as $\lambda_{data}=0.1$ and $\lambda_{cnn}=1$  based upon the results of a previous study using cyclic loss (21).

The entire training process were implemented in standard Python (v2.7, Python Software Foundation, Wilmington, Delaware). The network was designed using the Keras package (46) running Tensorflow computing backend (47) on a 64-bit Ubuntu Linux system. All training and evaluation was performed on a computer server with an Intel Xeon W3520 quad-core CPU, 32 GB DDR3 RAM, and one NVidia GeForce GTX 1080Ti graphic card with a total of 3584 CUDA cores and 11GB GDDR5 RAM.

\subsection{Sampling-Augmented Training Strategy}

\begin{figure}[h]
  \centering
  \includegraphics[width=0.8\linewidth]{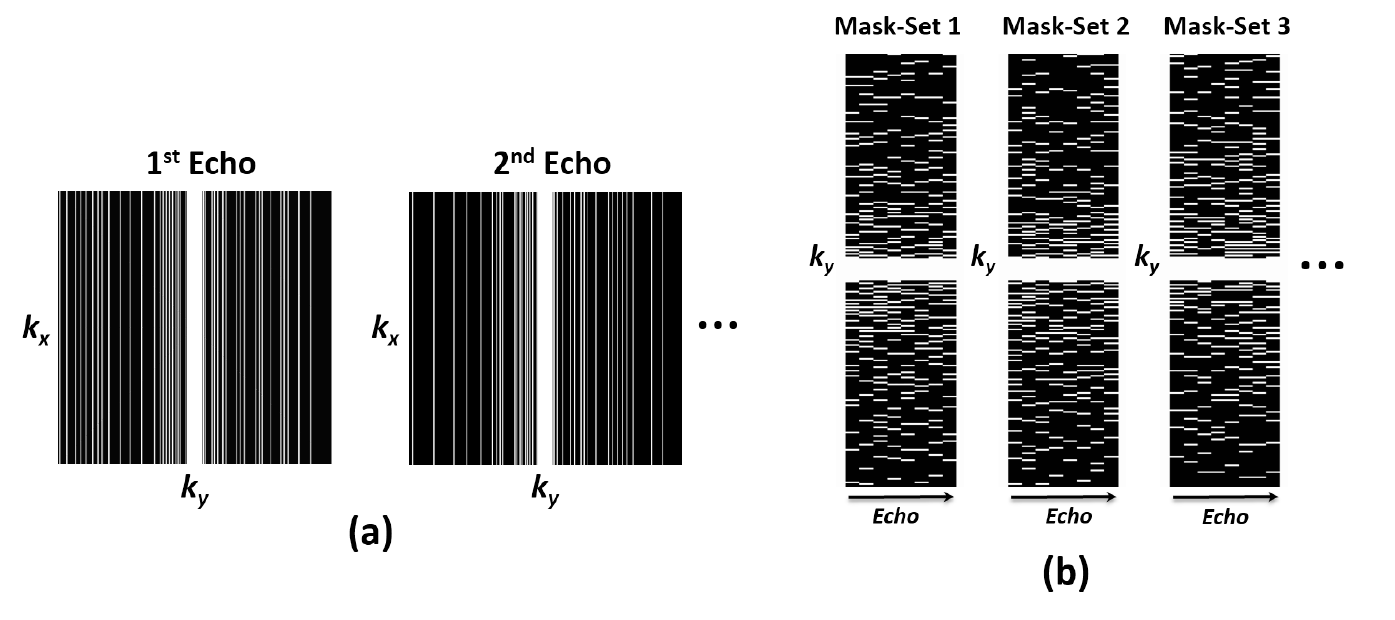}
  \caption{Schematic demonstration of the undersampling patterns used in the study. (a) Examples of the applied 1D variable-density random undersampled mask used for 1\textsuperscript{st} and 2\textsuperscript{nd} echo image. (b) Example sets of k\textsubscript{y}-t 1D variable-density random undersampling masks for eight echo times. The undersampling mask was varying along echo dimension and the mask set was randomized for each iteration during the network training to augment the training data.}
 \label{fig3}
\end{figure}

In order to improve the robustness of MANTIS against \textit{k}-space trajectory discrepancy between the training and testing datasets, a sampling-augmented training strategy was applied in our network training. Specifically, a sampling pattern library consisting of different sets of time-varying 1D variable-density random undersampling masks was first generated, as shown in Figure\ref{fig3}a. Each mask-set has eight different sampling masks (for introducing temporal incoherence) matching the number of echoes in our datasets, as shown in Figure\ref{fig3}b for one representative example. During the training process, a mask-set was randomly selected from this library for one training iteration, so that the network can learn a wide range of undersampling artifact structures during the training. It is our hypothesis that such a strategy can improve the robustness of the network against \textit{k}-space trajectory discrepancy, and thus the trained network can be used to reconstruct undersampled images acquired with different undersampling patterns. \\

For each undersampling pattern, 5\% of the \textit{k}-space center was fully sampled and two different acceleration rates (R) were investigated, including R=5 and R=8.

\subsection{Evaluation of Reconstruction Methods}

The performance of MANTIS was evaluated by comparing with 1) conventional sparsity-based reconstruction approaches exploring low-rank property of the multiecho image-series globally (referred to as GLR hereafter) and locally (referred to as LLR hereafter) respectively; and 2) direct CNN training using only the loss 2 component in Figure 1 (referred to as “CNN-Only” hereafter).

The GLR reconstruction aimed to solve the following objective function: 

\begin{equation} \label{eq10}
\tilde x = \arg \;\mathop {\min }\limits_x \left( {\frac{1}{2}\left\| {Ex - d} \right\|_2^2 + {\lambda _{GLR}}{{\left\| {Tx} \right\|}_*}} \right)
\end{equation}

Here, $E$ is an operator as previously shown in Eq.\ref{eq3}; $x$ is the multiecho image-series to be reconstructed with a size of $n_x \times n_y \times t$; $d$ is the corresponding undersampling \textit{k}-space with a size of $n_x \times n_y \times t$; $T$ is an operator to transform the 2D+time image-series to a Casorati matrix, in which each column corresponds to image pixels from each echo point. ${\left\| {Tx} \right\|_*}$ represents the nuclear norm of $Tx$, which is calculated as the sum of the singular values of $Tx$.

The LLR reconstruction was formulated as the following cost function: 

\begin{equation} \label{eq11}
\tilde x = \arg \;\mathop {\min }\limits_x \left( {\frac{1}{2}\left\| {Ex - d} \right\|_2^2 + {\lambda _{LLR}}\sum\limits_{b \in \Omega } {{{\left\| {{T_b}x} \right\|}_*}} } \right)
\end{equation}

In contrast to the GLR reconstruction, the LLR reconstruction aimed to exploit low-rank property in a local patch for improved reconstruction performance(13). Here, each 2D+time images $x$ can be partitioned into a set $\Omega$ of image blocks, in which each block represented a dynamic image-series with a smaller size (e.g., $b_x \times b_y \times t$). $T_b$ was then an operator to transform images from each block to a Casorati matrix as in the GLR reconstruction.

Both the GLR and LLR reconstructions were implemented using an iterative soft thresholding (ISTA) algorithm adapted from that proposed by Zhang et al in Ref  (13). The regularization parameters were empirically selected for each reconstruction type separately and were fixed for all reconstructions. For all the datasets, the GLR reconstruction was performed with a total of 50 iterations. Following the reconstruction implemented in Ref  (13) , the LLR reconstruction was initialized with GLR reconstruction for 20 iterations first, then followed by LLR reconstruction with reduced regularization weight for another 30 iterations. The block size was selected as 8$\times$8 was selected and the reconstruction was implemented with overlapping blocks to minimize blocky effect. For each image dataset, a T2 map was estimated by fitting Eq.\ref{eq2} from the reconstructed images on a pixel-by-pixel basis as described above.

The normalized Root Mean Squared Error (nRMSE) and Structural SIMilarity (SSIM) index, calculated with respect to the reference, were used to assess the overall reconstructed image errors. The nRMSE is defined as:

\begin{equation} \label{eq12}
nRMSE = \frac{{{{\left\| {{{\rm{T}}_{\rm{2}}} - {{{\rm{\tilde T}}}_{\rm{2}}}} \right\|}_{2,\Phi }}}}{{{{\left\| {{{\rm{T}}_{\rm{2}}}} \right\|}_{2,\Phi }}}}
\end{equation}

where ${{\rm{T}}_{\rm{2}}}$ and ${{\rm{\tilde T}}_{\rm{2}}}$ were estimated from the reference and accelerated data, respectively, and ${\left\|  \cdot  \right\|_{2,\Phi }}$ denoted the $l_2$ norm measured over the knee region $\Phi$.

\subsection{Regional T2 Analysis}

Region-of-interest (ROI) analysis was performed to compare the mean T2 values of the cartilage and meniscus from GLR, LLR, CNN-Only, and MANTIS at both R=5 and R=8 with the reference T2 values from the fully sampled images. Manual segmentation of the patellar, femoral and tibial cartilage and meniscus of the knee in the 10 testing patient datasets was performed by a research scientist with eight years of experience in medical image segmentation under the supervision of an experienced musculoskeletal radiologist using the first echo image and the reference T2 map. In addition, the cartilage from all the 10 patients was further divided into deep and superficial halves for sub-regional T2 analysis. Agreement was assessed using the Bland-Altman analysis.  Differences between the reconstructed and reference T2 values were assessed using the non-parametric Wilcoxon signed-rank test for the rank differences between paired measurements.  Statistical significance was defined as a \textit{p}-value less than 0.05.

\section{RESULTS}
The average time for the total network training process was about 19.4 hours. Following network training, the average time for reconstructing T2 maps in all image slices was about 8.1 seconds for each patient dataset.

\subsection{Evaluation of Reconstruction Methods}

\begin{table}[t]
  \caption{nRMSE and SSIM between the reference T2 maps estimated from the fully sampled images and the reconstructed T2 maps estimated using undersampling patterns.  Results were averaged over the 10 test patient datasets and represent mean value $\pm$ standard deviation. MANTIS achieved the highest reconstruction performance with the smallest errors at both R=5 and 8.}
   \label{tb1}
  \centering
  \begin{tabular}{llllll}
    \toprule
    & \multicolumn{2}{c}{R=5} & & \multicolumn{2}{c}{R=8}\\
    \cmidrule{2-3} \cmidrule{5-6}
    Methods   &  nRMSE(\%)      &  SSIM(\%)   & &  nRMSE(\%)   &  SSIM(\%) \\
    \midrule
    GLR 	  & 13.5$\pm$4.3  & 72.5$\pm$3.7   & & 15.0$\pm$3.9  & 63.2$\pm$4.5  \\
    LLR       & 12.2$\pm$3.5  & 70.4$\pm$3.1   & & 13.9$\pm$3.5  & 59.2$\pm$3.3 \\
    CNN-Only  & 6.9$\pm$1.8  & 82.3$\pm$2.8   & & 8.5$\pm$2.5 & 78.0$\pm$2.5 \\
    MANTIS    & 6.1$\pm$1.5  & 86.2$\pm$1.9   & & 7.1$\pm$1.8  & 82.1$\pm$2.3 \\
    \bottomrule
  \end{tabular}
\end{table}

\begin{figure}[h]
  \centering
  \includegraphics[width=0.8\linewidth]{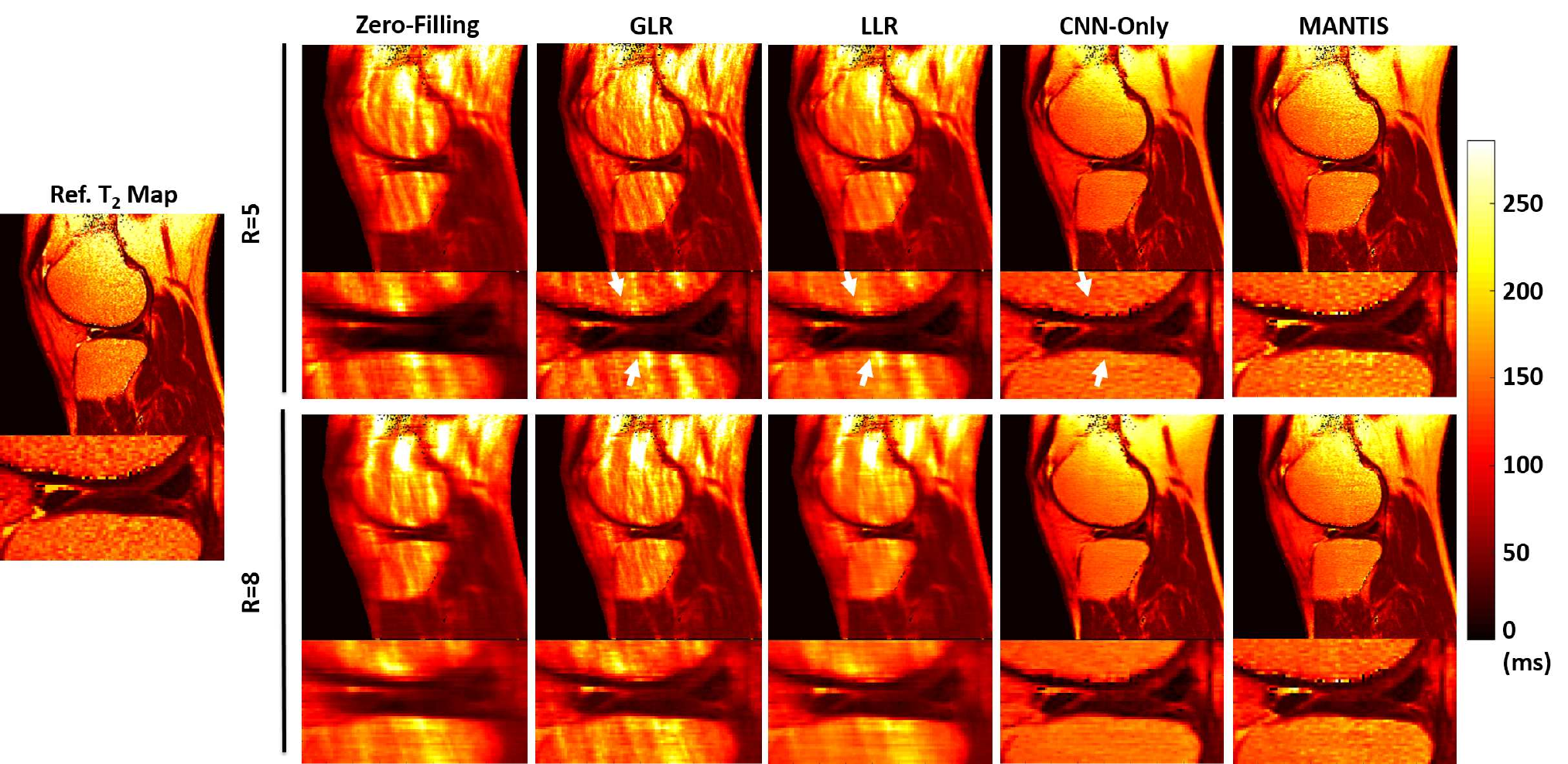}
  \caption{Representative examples of T2 maps estimated from the different reconstruction methods at R=5 (top row) and R=8 (bottom row), respectively. MANTIS generated nearly artifact-free T2 map with well-preserved sharpness and texture comparable to the reference T2 maps. The other methods generated suboptimal T2 maps with either reduced image sharpness or residual artifacts indicated by the white arrows.}
 \label{fig4}
\end{figure}

\begin{figure}[h]
  \centering
  \includegraphics[width=0.7\linewidth]{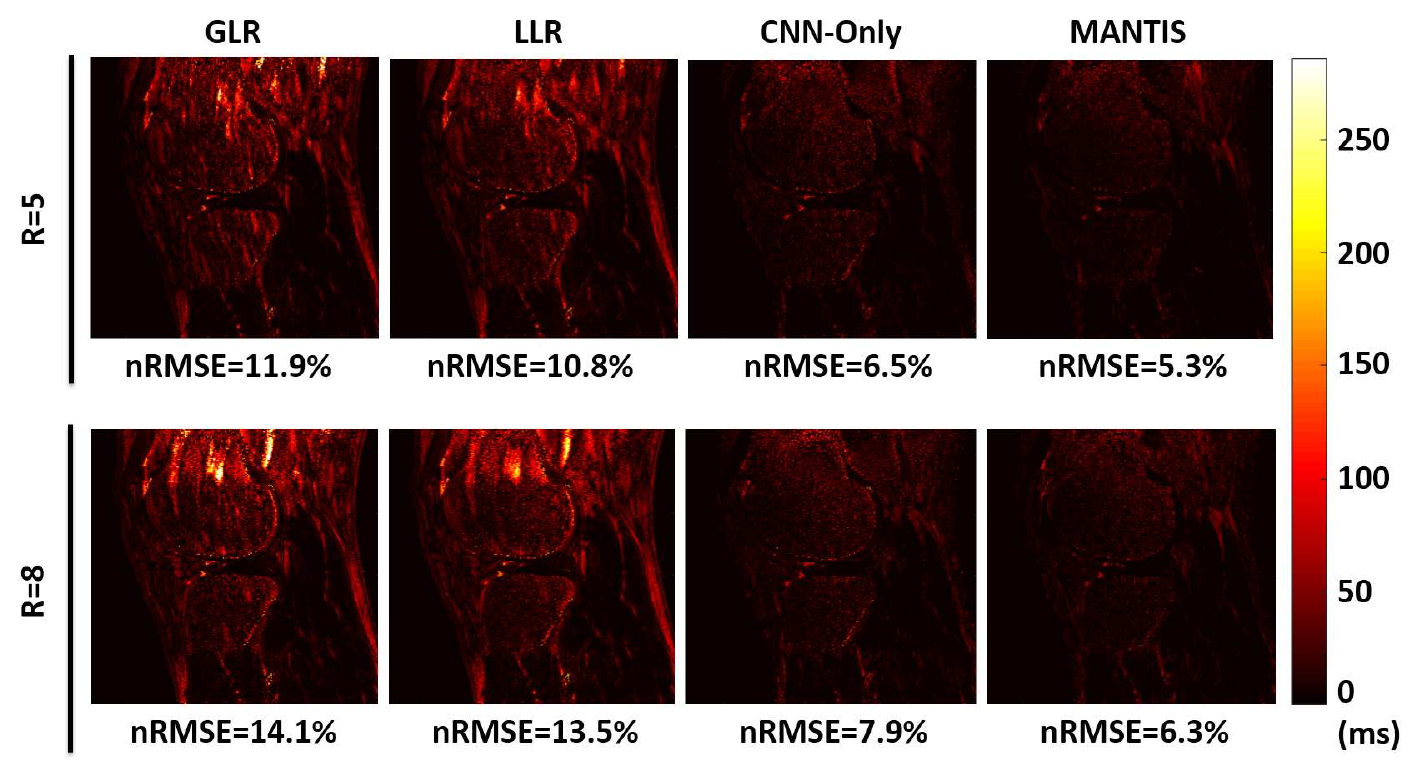}
  \caption{Residual error maps and  corresponding  nRMSE values from the  same patient shown in Figure\ref{fig4} which compares T2 maps estimated from the different reconstruction methods at R=5 (top row) and R=8 (bottom row), respectively.}
 \label{fig5}
\end{figure}

Table\ref{tb1} summarizes the mean nRMSE and SSIM values between the reference T2 maps and the reconstructed T2 maps averaged over all the 10 testing patient datasets. In general, The CNN-Only and MANTIS reconstruction methods were superior to the GLR and LLR reconstruction methods at both R=5 and R=8.  MANTIS yielded the smallest reconstruction errors and the highest similarity to the reference at both acceleration rates. Figure\ref{fig4} shows representative T2 maps estimated from different reconstruction methods at R=5 (top row) and R=8 (bottom row), respectively, for a symptomatic patient. The GLR reconstruction generated images with inferior image quality with noticeable artifacts in bone and fatty tissues. Although the LLR reconstruction improved overall image quality with reduced image artifacts, it led to a noticeable smooth appearance due to the exploitation of local sparsity at high acceleration. CNN-Only produced T2 maps with reduced image artifacts, but the sharpness and texture details of the reconstructed image were still suboptimal as indicated by the white arrows.  MANTIS generated nearly artifact-free T2 maps with well-preserved sharpness and texture comparable to the reference T2 maps. This qualitative observation was also confirmed by the corresponding residual error maps and nRMSE values from the same patient shown in Figure\ref{fig5}. The corresponding residual error maps are shown at the same scale to qualitatively compare the reconstruction performance.

\begin{figure}[h]
  \centering
  \includegraphics[width=0.8\linewidth]{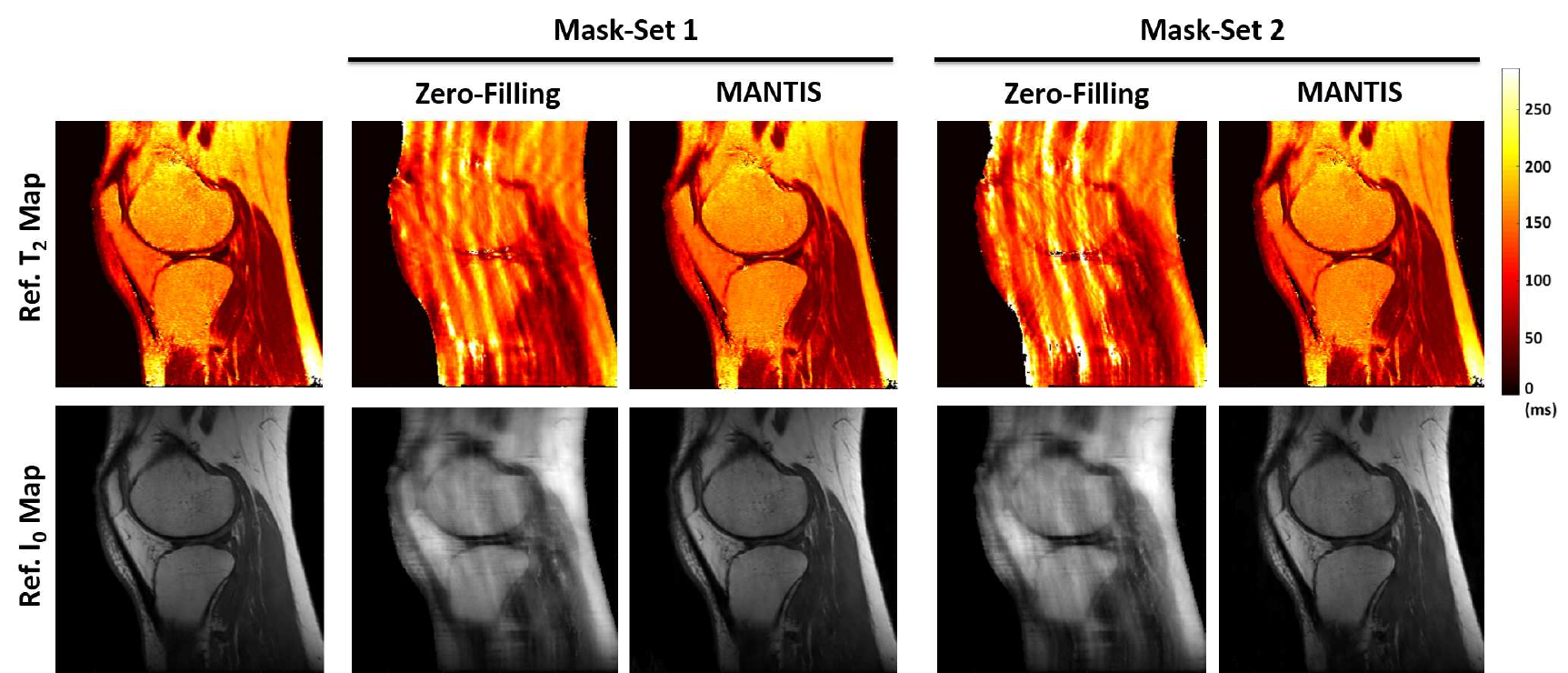}
  \caption{Comparison of T2 (top row) and I0 (bottom row) maps reconstructed using MANTIS at different undersampling masks for the healthy volunteer using raw \textit{k}-space data. Although the image artifacts arisen from the different undersampling masks are dramatically different, the MANTIS was able to remove the heterogeneous artifacts and provided nearly artifact-free T2 and I0 maps regardless of different undersampling masks.}
 \label{fig6}
\end{figure}

Figure\ref{fig6} shows the T2 map (top row) and proton density I0 map (bottom row) reconstructed from MANTIS in the volunteer with raw \textit{k}-space data with different \textit{k}-space undersampling patterns at R=5. Due to the applied different undersampling masks, there was a significant difference of image artifacts for the same image from the zero-filling reconstruction. However, regardless the difference of the undersampling masks, MANTIS achieved a great reconstruction performance for suppressing the heterogeneous image artifacts and maintaining image quality and sharpness that was comparable to the fully sampled reference T2 and I0 maps. Incorporation of additional incoherence at dynamic frame and training phase improved image quality and resulted in a robust MANTIS model. 

\subsection{Lesion Detection}
\begin{figure}[h]
  \centering
  \includegraphics[width=0.7\linewidth]{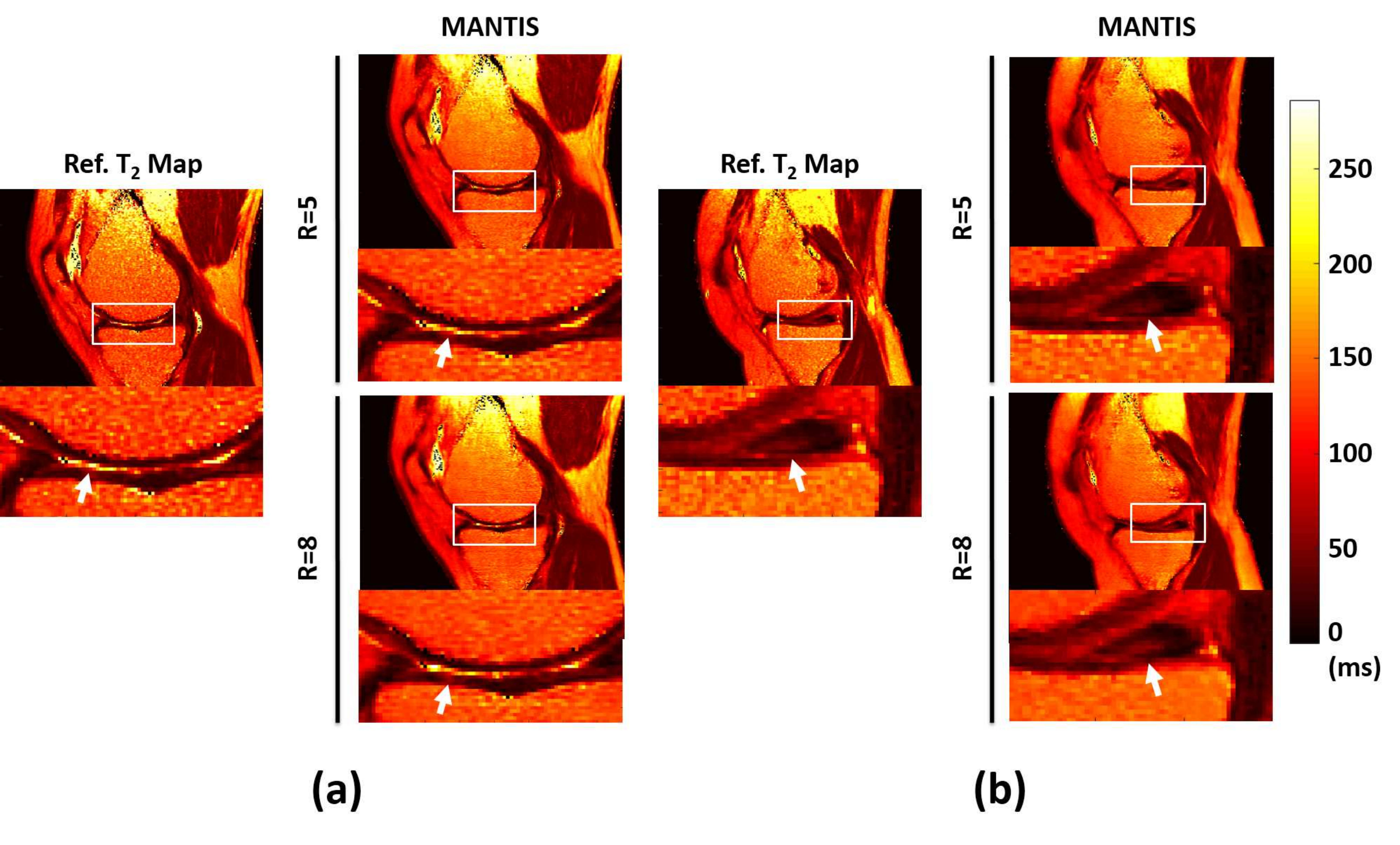}
  \caption{Two representative examples demonstrating the performance of MANTIS in cartilage and meniscus lesion detection. a) Results from a 67-year male patient with knee osteoarthritis and superficial cartilage degeneration on the medial femoral condyle and medial tibia plateau. b) Results from a 59-year male patient with a tear of the posterior horn of the medial meniscus.  MANTIS was able to reconstruct high quality T2 maps for clear identification of cartilage and meniscus lesions at both R=5 and R=8.}
 \label{fig7}
\end{figure}

Figure\ref{fig7} compares T2 maps between MANTIS and the reference in two representative patients. Figure\ref{fig7}a shows T2 maps from a 67-year male patient with knee osteoarthritis with superficial cartilage degeneration on the medial femoral condyle and medial tibia plateau. The morphologic abnormalities and increased T2 relaxation time in the superficial cartilage could be identified in the reconstructed T2 maps at R=5. While the overall reconstruction quality was slightly reduced at R=8, the high contrast abnormalities could still be reliably identified. Figure\ref{fig7}b shows T2 maps from a 59-year male patient with a tear of the posterior horn of the medial meniscus. The reference T2 maps shows a heterogeneous increase in T2 relaxation time at the center of the meniscus extending into the inferior articular surface, which could also be successfully captured on the reconstructed T2 maps using MANTIS at both R=5 and R=8.

\subsection{Regional Cartilage and Meniscus T2 Analysis}

\begin{table}[t]
  \caption{Regional cartilage and meniscus T2 analysis for the 10 testing patient datasets using the different reconstruction methods at acceleration rate R=5 and R=8. MANTIS provided mean T2 values that were closest to the reference T2 values for the patellar, femoral, and tibial cartilage, superficial and deep half cartilage, and meniscus.}
   \label{tb2}
  \centering
  \begin{tabular}{llllllll}
    \toprule
    & \multicolumn{6}{c}{T2 values(ms) at R=5} \\
    \cmidrule{2-7}
    Methods   &  Patellar & Femoral & Tibial & Superficial & Deep & Meniscus  \\
    \midrule
    GLR 	  & 55.0$\pm$3.2  & 54.7$\pm$2.8   & 48.7$\pm$5.9   & 66.0$\pm$3.8   & 39.6$\pm$2.3   & 37.3$\pm$2.8 \\
    LLR       & 52.4$\pm$3.4  & 53.1$\pm$2.8   & 47.1$\pm$6.2   & 63.6$\pm$3.9   & 38.2$\pm$2.3   & 34.7$\pm$3.1 \\
    CNN-Only  & 41.1$\pm$3.7  & 45.9$\pm$3.2   & 41.5$\pm$6.1   & 53.5$\pm$4.4   & 32.1$\pm$2.7   & 28.0$\pm$3.9 \\
    MANTIS    & 40.4$\pm$3.7  & 45.6$\pm$3.4   & 41.7$\pm$5.6   & 53.2$\pm$4.1   & 31.9$\pm$2.4   & 28.3$\pm$4.0 \\
    Reference & 39.6$\pm$3.5  & 46.0$\pm$3.4   & 42.5$\pm$5.5   & 53.4$\pm$3.8   & 32.0$\pm$2.3   & 27.5$\pm$4.0 \\
    \\
    & \multicolumn{6}{c}{T2 values(ms) at R=8} \\
    \cmidrule{2-7}
    GLR 	  & 57.4$\pm$3.1  & 56.8$\pm$2.5   & 49.6$\pm$5.5   & 68.2$\pm$3.5   & 40.9$\pm$2.1   & 38.0$\pm$3.4 \\
    LLR       & 52.7$\pm$3.3  & 53.5$\pm$2.8   & 47.4$\pm$6.1   & 64.0$\pm$3.9   & 38.4$\pm$2.3   & 34.9$\pm$3.1 \\
    CNN-Only  & 42.6$\pm$4.4  & 44.5$\pm$3.4   & 39.1$\pm$7.4   & 52.9$\pm$5.0   & 31.7$\pm$3.0   & 28.4$\pm$3.9 \\
    MANTIS    & 41.0$\pm$3.9  & 45.5$\pm$3.6   & 41.1$\pm$5.8   & 53.3$\pm$4.3   & 32.0$\pm$2.8   & 28.7$\pm$4.5 \\
    Reference & 39.6$\pm$3.5  & 46.0$\pm$3.4   & 42.5$\pm$5.5   & 53.4$\pm$3.8   & 32.0$\pm$2.3   & 27.5$\pm$4.0 \\
    
    \bottomrule
  \end{tabular}
\end{table}

Table\ref{tb2} summarizes the results of the regional T2 analysis for the 10 testing patient datasets at different acceleration rates. MANTIS provided mean T2 values that were closest to the reference in the patellar, femoral, and tibial cartilage, the deep and superficial half cartilage, and the meniscus. There were significant differences in the estimated T2 values between both the GLR and LLR method and the reference (\textit{p}<0.001) at both R=5 and R=8. In contrast, there was no significant difference between the deep learning-based methods and the reference, with \textit{p}=0.34 (R=5) and \textit{p}=0.29 (R=8) for CNN-Only and \textit{p}=0.90 (R=5) and \textit{p}=0.57 (R=8) for MANTIS, respectively.

\begin{figure}[h] 
  \centering
  \includegraphics[width=0.8\linewidth]{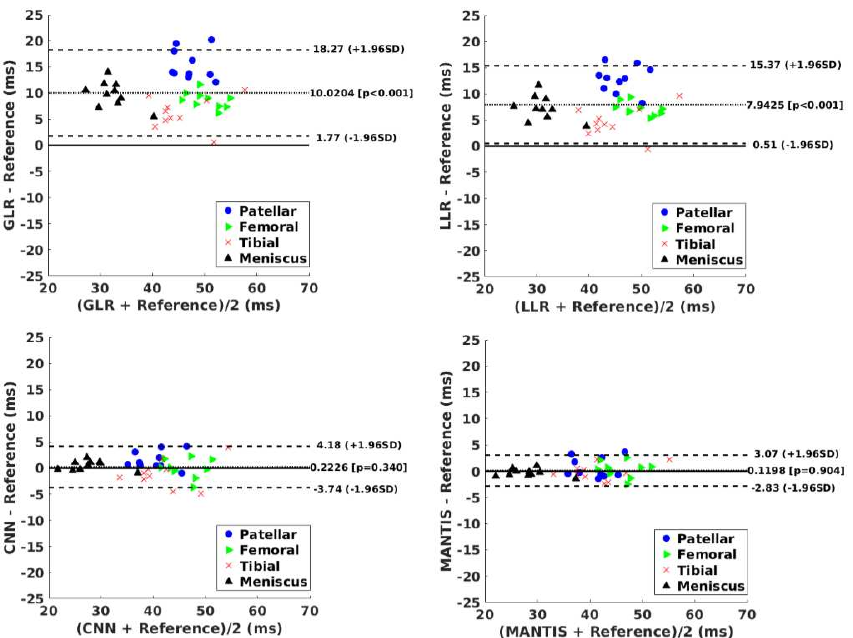}
  \caption{The Bland-Altman analysis for agreement of the regional cartilage and meniscus T2 values obtained using the reference T2 maps and the T2 maps estimated using the different reconstruction methods at R=5.}
\label{fig8}
\end{figure}

\begin{figure}[h] 
  \centering
  \includegraphics[width=0.8\linewidth]{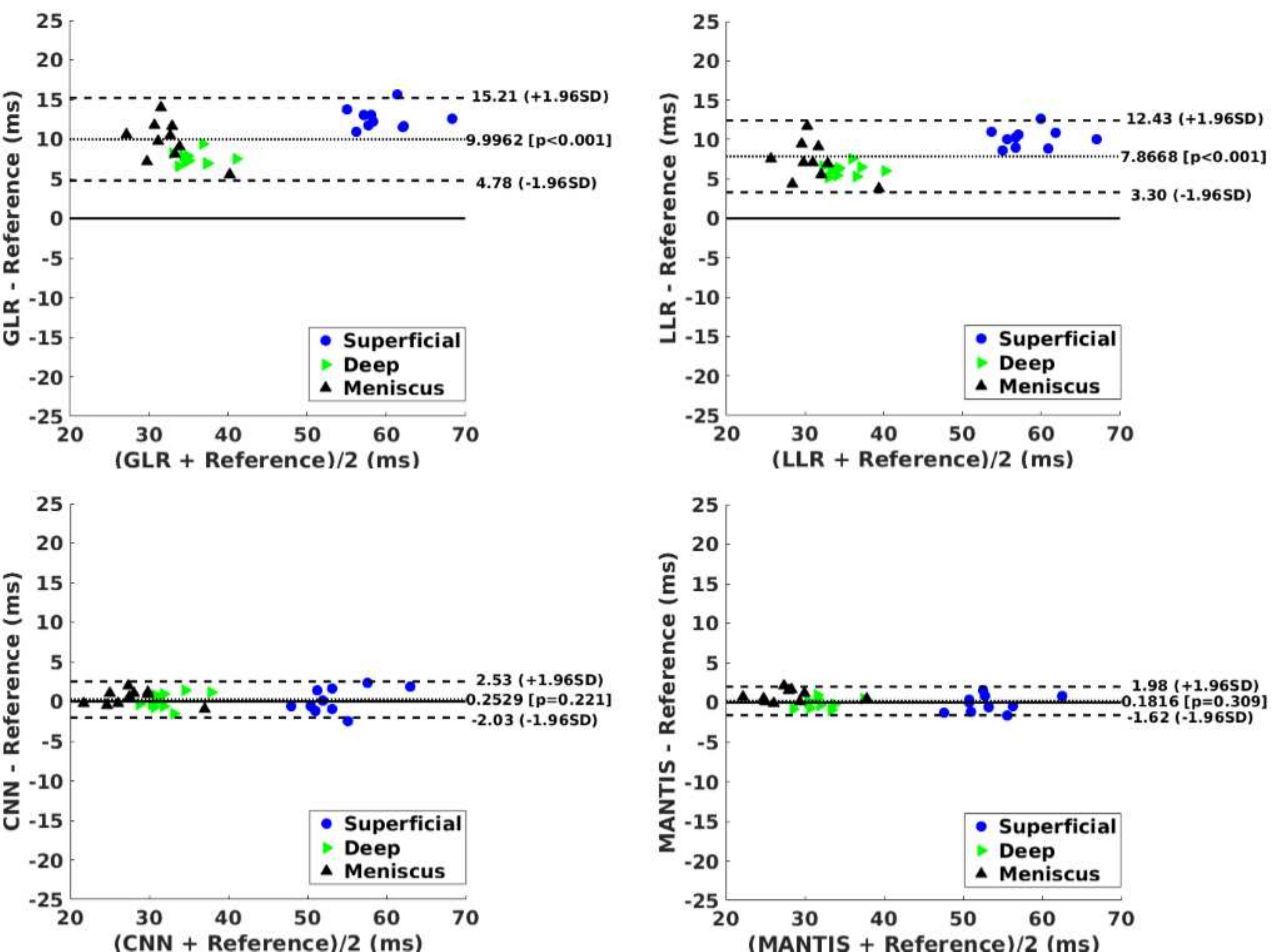}
  \caption{The Bland-Altman analysis for agreement of the sub-regional cartilage (superficial and deep halves) and meniscus T2 values obtained using the reference T2 maps and the T2 maps estimated using the different reconstruction methods at R=5.}
\label{fig9}
\end{figure}

Figures\ref{fig8} shows the Bland-Altman plots comparing the reference T2 maps with reconstructed T2 maps in the cartilage and meniscus at R=5.  Compared to GLR and LLR, both CNN-Only and MANTIS achieved greater agreement with the reference T2 values for cartilage and meniscus, as indicated by the narrower limits of agreements at $\pm$1.96$\times$standard deviation of the mean differences in the plots. MANTIS achieved further improved agreement compared to CNN-Only. The Bland-Altman plots for the sub-regional superficial and deep cartilage is shown in Figure\ref{fig9}. A similar observation that MANTIS archived the best agreement with the reference compared to other approaches was noted. 

\section{DISCUSSION}
In this work, a deep learning-based reconstruction framework called MANTIS was proposed for efficient T2 mapping in the knee joint. To the best of our knowledge, MANTIS is the one of the pilot methods described for direct parameter mapping from undersampled images using deep learning. The MANTIS pipeline consists of several key components. As the core component of the reconstruction framework, a convolutional encoder-decoder network is designed for directly converting a series of undersampled MR images to T2 maps using supervised end-to-end CNN mapping. Meanwhile, time-varying random \textit{k}-space sampling is employed to ensure good performance of the CNN mapping by taking advantage of the resulting incoherent undersampling behavior. Furthermore, a signal model is incorporated into MANTIS to enforce data and model consistency for optimal reconstruction performance. The entire reconstruction framework is formulated as a regularized reconstruction approach, in line with a conventional constraint reconstruction pipeline, where the end-to-end CNN mapping serves as a regularizer to the data/model consistency component.

In our study, MANTIS enabled up to eight-fold acceleration with acceptable reconstruction performance and accurate T2 estimations of the cartilage and meniscus with respect to the fully sampled reference. In addition to global nRMSE and SSIM assessment, regional T2 analysis was also conducted to further confirm that MANTIS produced accurate T2 values in the cartilage and meniscus.  Moreover, MANTIS was proven to be highly time efficient.  Although the network was trained up to 20 hours, the training only needed to be performed once for a particular application, and the trained MANTIS network can be deployed to generate MR T2 maps directly within seconds. Such a high runtime efficiency in MANTIS holds greater promise towards translation for routine clinical use, compared to many existing iterative reconstruction approaches that normally take up to hours.

In this work, MANTIS was evaluated with a 1D variable-density undersampling scheme to allows good performance in the CNN mapping step. This was inspired by the compressed sensing theory and the MRF framework, in which reduced correlations of aliasing behavior arising from the incoherent sampling pattern is more beneficial for distinguishing subtle anatomical structures from structural artifacts. Furthermore, the strategy of using randomized undersampling patterns to augment CNN training was shown to increase the robustness of the proposed framework against different undersampling artifacts (Figure\ref{fig6}). This feature is important, since the robustness of learning-based image reconstruction framework against \textit{k}-space trajectory discrepancy between the training and testing datasets is of great interest (48). In addition to random Cartesian sampling, we expect that non-Cartesian sampling schemes, such as radial or spiral, are also well-suited for the MANTIS framework. The extension of MANTIS into non-Cartesian sampling is of high interest and can further improve the mapping performance and clinical applicability of the technique. Additional work is currently under planning to evaluate the feasibility of combining MANTIS with non-Cartesian sampling.

MANTIS was compared with conventional GLR and LLR reconstruction methods, both of which have previously been demonstrated with good performance in accelerated T2 mapping (13). MANTIS significantly outperformed both approaches at both R=5 and R=8. The suboptimal reconstruction performance of GLR and LLR can be attributed to the fact that a high acceleration rate (up to R=8) was beyond the sparsity level of the T2 mapping image-series (eight echo images only with limited temporal correlations) in our study, particularly with a single-coil setup. In addition, MANTIS was also compared with standard CNN mapping without the data and model consistency components. MANTIS achieved superior performance in the nRMSE and SSIM analyses and regional T2 assessment. 

Although our current study demonstrated the feasibility of MANTIS for accelerated T2 mapping of the knee, the approach can be extended to combine other models for various applications. For example, MANTIS can be used for mapping apparent diffusion coefficients (ADC) or intravoxel incoherent motion (IVIM) parameters in diffusion imaging and for direct estimation of pharmacokinetic parameters in perfusion imaging. The accuracy of the estimation is dependent on the model employed to generate the reference maps for the network training. We elected to use a simple mono-exponential decay model for reconstructing the T2 maps for the current feasibility study. Although this mono-exponential model has been shown to be suboptimal for a multi-echo spin-echo sequence (49,50), improvement can be obtained by substituting this simple model with more advanced signal models. For example, Ben-Eliezer et.al. have shown that an echo modulation curve algorithm, which is based on Bloch simulations to model the exact signal evolution in a multi-echo spin-echo sequence, can be used to produce more reliable T2 parameter estimations (49). Recently, another study has also demonstrated the feasibility of using a spin-echo sequence with different echo times, the current gold standard for T2 mapping, to generate accurate T2 maps for subsequent network training (51). Although the method in (51) can be extremely time-consuming, the simulation-based approach (49) is ideal for incorporation into the MANTIS framework to further improve the accuracy of T2 parameter estimation. This combination is of great interest and is currently underway. Furthermore, MANTIS can also be used for efficient estimation of both T1 and T2 simultaneously using the MRF framework. One pilot study has demonstrated the feasibility of mapping images generated from MRF to parameter maps using an end-to-end fully connected neural network (31). Such a mapping strategy can also be combined with the MANTIS framework to promote further acceleration for parameter mapping.  

Our current study has several limitations. First, the feasibility of MANTIS was demonstrated in a single-coil scenario only.  The extension of MANTIS for multi-coil use is entirely possible provided that appropriate and adequate multi-coil training dataset, better GPU architecture, and increased GPU memory are available. Second, the current study used a U-Net structure and did not compare the usefulness of U-Net with other end-to-end CNN mapping structures. This selection was based on prior studies that have justified the performance of U-Net for a wide range of image reconstruction and analysis tasks (24,27,39). However, it would still be interesting to compare the U-Net with newly developed deep learning networks (20,25) for the implementation of the MANTIS framework. Third, the parameters used in the network training process were selected based on heuristic information from previous studies. This is similar to the parameter tuning in constrained reconstruction, where a regularization weight is normally tailored to a specific application. The selection of parameter in MANTIS is also dependent on the application and the quality and the number of training datasets, like other learning-based methods (20,21). Future work involving a comprehensive parameter optimization strategy is needed to validate the optimized performance of the network and the sensitivity of reconstruction results to these parameters.

\section{CONCLUSIONS}
Our study demonstrated that the proposed MANTIS framework, with a combination of end-to-end CNN mapping, signal model-augmented data consistency, and incoherent \textit{k}-space sampling, represents a promising approach for efficient T2 mapping. MANTIS can potentially be extended to other types of parameter mapping such as T1 relaxation time, diffusion, and perfusion with appropriate models or to a combination of these in the context of MRF as long as a sufficient number of training datasets are available.

\section*{REFERENCES}
\medskip
\small
[1] Pruessmann KP, Weiger M, Scheidegger MB, Boesiger P. {\it SENSE: Sensitivity Encoding for Fast MRI.} Magn. Reson. Med. 1999;42:952–962.

[2] Sodickson DK, Manning WJ. {\it Simultaneous acquisition of spatial harmonics (SMASH): fast imaging with radiofrequency coil arrays.} Magn. Reson. Med. [Internet] 1997;38:591–603.

[3] Griswold MA, Jakob PM, Heidemann RM, Nittka M, Jellus V, Wang J, Kiefer B, Haase A. {\it Generalized Autocalibrating Partially Parallel Acquisitions (GRAPPA).} Magn. Reson. Med. 2002;47:1202–1210. doi: 10.1002/mrm.10171.

[4] Lustig M, Donoho D, Pauly JM. {\it Sparse MRI: The application of compressed sensing for rapid MR imaging.} Magn Reson Med [Internet] 2007;58:1182–1195. doi: 10.1002/mrm.21391.

[5] Lustig M, Pauly JM. {\it SPIRiT: Iterative self-consistent parallel imaging reconstruction from arbitrary \textit{k}-space.} Magn. Reson. Med. [Internet] 2010;64:457–471. doi: 10.1002/mrm.22428.

[6] Otazo R, Kim D, Axel L, Sodickson DK. {\it Combination of compressed sensing and parallel imaging for highly accelerated first-pass cardiac perfusion MRI.} Magn. Reson. Med. [Internet] 2010;64:767–776. doi: 10.1002/mrm.22463.

[7] Doneva M, Börnert P, Eggers H, Stehning C, Sénégas J, Mertins A. {\it Compressed sensing reconstruction for magnetic resonance parameter mapping.} Magn. Reson. Med. [Internet] 2010;64:1114–1120. doi: 10.1002/mrm.22483.

[8] Block KT, Uecker M, Frahm J. {\it Model-based iterative reconstruction for radial fast spin-echo MRI.} IEEE Trans. Med. Imaging [Internet] 2009;28:1759–69. doi: 10.1109/TMI.2009.2023119.

[9] Huang C, Graff CG, Clarkson EW, Bilgin A, Altbach MI. {\it T2 mapping from highly undersampled data by reconstruction of principal component coefficient maps using compressed sensing.} Magn. Reson. Med. [Internet] 2012;67:1355–66. doi: 10.1002/mrm.23128.

[10] Petzschner FH, Ponce IP, Blaimer M, Jakob PM, Breuer FA. {\it Fast MR parameter mapping using k-t principal component analysis.} Magn. Reson. Med. [Internet] 2011;66:706–16. doi: 10.1002/mrm.22826.

[11] Feng L, Otazo R, Jung H, Jensen JH, Ye JC, Sodickson DK, Kim D. {\it Accelerated cardiac T2 mapping using breath-hold multiecho fast spin-echo pulse sequence with k-t FOCUSS.} Magn. Reson. Med. [Internet] 2011;65:1661–1669. doi: 10.1002/mrm.22756.

[12] Velikina J V., Alexander AL, Samsonov A. {\it Accelerating MR parameter mapping using sparsity-promoting regularization in parametric dimension.} Magn. Reson. Med. 2013;70:1263–1273. doi: 10.1002/mrm.24577.

[13] Zhang T, Pauly JM, Levesque IR. {\it Accelerating parameter mapping with a locally low rank constraint.} Magn. Reson. Med. [Internet] 2015;73:655–661. doi: 10.1002/mrm.25161.

[14] Sumpf TJ, Uecker M, Boretius S, Frahm J. {\it Model-based nonlinear inverse reconstruction for T2 mapping using highly undersampled spin-echo MRI.} J. Magn. Reson. Imaging [Internet] 2011;34:420–428. doi: 10.1002/jmri.22634.

[15] Wang X, Roeloffs V, Klosowski J, Tan Z, Voit D, Uecker M, Frahm J. {\it Model-based T1mapping with sparsity constraints using single-shot inversion-recovery radial FLASH.} Magn. Reson. Med. [Internet] 2018;79:730–740. doi: 10.1002/mrm.26726.

[16] Ma D, Gulani V, Seiberlich N, Liu K, Sunshine JL, Duerk JL, Griswold MA. {\it Magnetic resonance fingerprinting.} Nature [Internet] 2013;495:187–192. doi: 10.1038/nature11971.

[17] Panda A, Mehta BB, Coppo S, Jiang Y, Ma D, Seiberlich N, Griswold MA, Gulani V. {\it Magnetic Resonance Fingerprinting- An Overview.} Curr. Opin. Biomed. Eng. [Internet] 2017;3:56–66. doi: 10.1016/j.cobme.2017.11.001.

[18] Hammernik K, Klatzer T, Kobler E, Recht MP, Sodickson DK, Pock T, Knoll F. {\it Learning a Variational Network for Reconstruction of Accelerated MRI Data.} Magn. Reson. Med. [Internet] 2017;79:3055–3071. doi: 10.1002/mrm.26977.

[19] Liu F, Samsonov A. {\it Data-Cycle-Consistent Adversarial Networks for High-Quality Reconstruction of Undersampled MRI Data.} In: the ISMRM Machine Learning Workshop. ; 2018.

[20] Quan TM, Nguyen-Duc T, Jeong W-K. {\it Compressed Sensing MRI Reconstruction using a Generative Adversarial Network with a Cyclic Loss.} IEEE Trans. Med. Imaging [Internet] 2017;37:1488–1497. doi: 10.1109/TMI.2018.2820120.

[21] Mardani M, Gong E, Cheng JY, Vasanawala SS, Zaharchuk G, Xing L, Pauly JM. {\it Deep Generative Adversarial Neural Networks for Compressive Sensing (GANCS) MRI.} IEEE Trans. Med. Imaging [Internet] 2018:1–1. doi: 10.1109/TMI.2018.2858752.

[22] Wang S, Su Z, Ying L, Peng X, Zhu S, Liang F, Feng D, Liang D. {\it Accelerating magnetic resonance imaging via deep learning.} In: 2016 IEEE 13th International Symposium on Biomedical Imaging (ISBI). IEEE; 2016. pp. 514–517. doi: 10.1109/ISBI.2016.7493320.

[23] Schlemper J, Caballero J, Hajnal J V., Price A, Rueckert D. {\it A Deep Cascade of Convolutional Neural Networks for Dynamic MR Image Reconstruction.} IEEE Trans. Med. Imaging [Internet] 2017:1–1. doi: 10.1007/978-3-319-59050-9\_51.

[24] Han Y, Yoo J, Kim HH, Shin HJ, Sung K, Ye JC. {\it Deep learning with domain adaptation for accelerated projection-reconstruction MR.} Magn. Reson. Med. [Internet] 2018;80:1189–1205. doi: 10.1002/mrm.27106.

[25] Eo T, Jun Y, Kim T, Jang J, Lee HJ, Hwang D. {\it KIKI-net: Cross-domain convolutional neural networks for reconstructing undersampled magnetic resonance images.} Magn. Reson. Med. [Internet] 2018. doi: 10.1002/mrm.27201.

[26] Lv J, Chen K, Yang M, Zhang J, Wang X. {\it Reconstruction of undersampled radial free-breathing 3D abdominal MRI using stacked convolutional auto-encoders.} Med. Phys. [Internet] 2018;45:2023–2032. doi: 10.1002/mp.12870.

[27] Kim KH, Do W-J, Park S-H. {\it Improving resolution of MR images with an adversarial network incorporating images with different contrast.} Med. Phys. [Internet] 2018;45:3120–3131. doi: 10.1002/mp.12945.

[28] Zhu B, Liu JZ, Cauley SF, Rosen BR, Rosen MS. {\it Image reconstruction by domain-transform manifold learning.} Nature [Internet] 2018;555:487–492. doi: 10.1038/nature25988.

[29] Golkov V, Dosovitskiy A, Sperl JI, et al. {\it q-Space Deep Learning: Twelve-Fold Shorter and Model-Free Diffusion MRI Scans.} IEEE Trans. Med. Imaging [Internet] 2016;35:1344–1351. doi: 10.1109/TMI.2016.2551324.

[30] Cai C, Wang C, Zeng Y, Cai S, Liang D, Wu Y, Chen Z, Ding X, Zhong J. {\it Single-shot T2 mapping using overlapping-echo detachment planar imaging and a deep convolutional neural network.} Magn. Reson. Med. [Internet] 2018:1–13. doi: 10.1002/mrm.27205.

[31] Cohen O, Zhu B, Rosen MS. {\it MR fingerprinting Deep RecOnstruction NEtwork (DRONE).} Magn. Reson. Med. [Internet] 2018;80:885–894. doi: 10.1002/mrm.27198.

[32] Zhu J-Y, Park T, Isola P, Efros AA. {\it Unpaired Image-to-Image Translation Using Cycle-Consistent Adversarial Networks.} In: 2017 IEEE International Conference on Computer Vision (ICCV). Vol. 2017–Octob. IEEE; 2017. pp. 2242–2251. doi: 10.1109/ICCV.2017.244.

[33] Klaas P. Pruessmann. {\it Advances in Sensitivity Encoding With Arbitrary in\textit{k}-space Trajectories.} 2001;651:638–651.

[34] Liu F, Zhou Z, Jang H, Samsonov A, Zhao G, Kijowski R. {\it Deep Convolutional Neural Network and 3D Deformable Approach for Tissue Segmentation in Musculoskeletal Magnetic Resonance Imaging.} Magn. Reson. Med. [Internet] 2017:DOI: 10.1002/mrm.26841. doi: 10.1002/mrm.26841.

[35] Zhou Z, Zhao G, Kijowski R, Liu F. {\it Deep Convolutional Neural Network for Segmentation of Knee Joint Anatomy.} Magn. Reson. Med. 2018:doi:10.1002/mrm.27229.

[36] Zhao G, Liu F, Oler JA, Meyerand ME, Kalin NH, Birn RM. {\it Bayesian convolutional neural network based MRI brain extraction on nonhuman primates.} Neuroimage [Internet] 2018;175:32–44. doi: 10.1016/j.neuroimage.2018.03.065.

[37] Liu F, Jang H, Kijowski R, Bradshaw T, McMillan AB. {\it Deep Learning MR Imaging–based Attenuation Correction for PET/MR Imaging.} Radiology [Internet] 2017:170700. doi: 10.1148/radiol.2017170700.

[38] Jang H, Liu F, Zhao G, Bradshaw T, McMillan AB. {\it Technical Note: Deep learning based MRAC using rapid ultra-short echo time imaging.} Med. Phys. [Internet] 2018:In-press. doi: 10.1002/mp.12964.

[39] Gong E, Pauly JM, Wintermark M, Zaharchuk G. {\it Deep learning enables reduced gadolinium dose for contrast-enhanced brain MRI.} J. Magn. Reson. Imaging [Internet] 2018. doi: 10.1002/jmri.25970.

[40] Chaudhari AS, Fang Z, Kogan F, Wood J, Stevens KJ, Gibbons EK, Lee JH, Gold GE, Hargreaves BA. {\it Super-resolution musculoskeletal MRI using deep learning.} Magn. Reson. Med. [Internet] 2018. doi: 10.1002/mrm.27178.

[41] Leynes AP, Yang J, Wiesinger F, Kaushik SS, Shanbhag DD, Seo Y, Hope TA, Larson PEZ. {\it Direct PseudoCT Generation for Pelvis PET/MRI Attenuation Correction using Deep Convolutional Neural Networks with Multi-parametric MRI: Zero Echo-time and Dixon Deep pseudoCT (ZeDD-CT).} J. Nucl. Med. [Internet] 2017:jnumed.117.198051. doi: 10.2967/jnumed.117.198051.

[42] Ronneberger O, Fischer P, Brox T. {\it U-Net: Convolutional Networks for Biomedical Image Segmentation.} In: Navab N, Hornegger J, Wells WM, Frangi AF, editors. Medical Image Computing and Computer-Assisted Intervention -- MICCAI 2015: 18th International Conference, Munich, Germany, October 5-9, 2015, Proceedings, Part III. Cham: Springer International Publishing; 2015. pp. 234–241. doi: 10.1007/978-3-319-24574-4\_28.

[43] Liu F, Kijowski R. {\it Assessment of different fitting methods for in-vivo bi-component T2(*) analysis of human patellar tendon in magnetic resonance imaging.} Muscles. Ligaments Tendons J. [Internet] 2017;7:163–172. doi: 10.11138/mltj/2017.7.1.163.

[44] He K, Zhang X, Ren S, Sun J. {\it Delving Deep into Rectifiers: Surpassing Human-Level Performance on ImageNet Classification.} ArXiv e-prints [Internet] 2015;1502.

[45] Kingma DP, Ba J. {\it Adam: A Method for Stochastic Optimization.} ArXiv e-prints [Internet] 2014.

[46] François Chollet. Keras. GitHub 2015:https://github.com/fchollet/keras.

[47] Abadi M, Agarwal A, Barham P, et al. {\it TensorFlow: Large-Scale Machine Learning on Heterogeneous Distributed Systems.} ArXiv e-prints [Internet] 2016. doi: 10.1109/TIP.2003.819861.

[48] Knoll F, Hammernik K, Kobler E, Pock T, Recht MP, Sodickson DK. {\it Assessment of the generalization of learned image reconstruction and the potential for transfer learning.} Magn. Reson. Med. [Internet] 2018. doi: 10.1002/mrm.27355.

[49] Ben-Eliezer N, Sodickson DK, Block KT. {\it Rapid and accurate T2 mapping from multi-spin-echo data using Bloch-simulation-based reconstruction.} Magn. Reson. Med. [Internet] 2015;73:809–817. doi: 10.1002/mrm.25156.

[50] Ben-Eliezer N, Sodickson DK, Shepherd T, Wiggins GC, Block KT. {\it Accelerated and motion-robust in vivo T2mapping from radially undersampled data using bloch-simulation-based iterative reconstruction.} Magn. Reson. Med. [Internet] 2016;75:1346–1354. doi: 10.1002/mrm.25558.

[51] Hilbert T, Thiran J-P, Meuli R, Kober T. {\it Quantitative Mapping by Data-Driven Signal-Model Learning.} In: the ISMRM 26th Annual Meeting. ; 2018. p. Abstract 777.

\end{document}